\documentclass[conference]{IEEEtran}
\IEEEoverridecommandlockouts
\usepackage{cite}
\usepackage{amsmath,amssymb,amsfonts}
\usepackage{algorithmic}
\usepackage{graphicx}
 \usepackage[colorlinks=true,linkcolor=blue, citecolor=blue]{hyperref}%
\usepackage{textcomp}
\usepackage{xcolor}
\def\BibTeX{{\rm B\kern-.05em{\sc i\kern-.025em b}\kern-.08em
    T\kern-.1667em\lower.7ex\hbox{E}\kern-.125emX}}

\begin{document}

\title{Navigating Healthcare Insights: A Bird's Eye View of Explainability with Knowledge Graphs\\
}

\author{\IEEEauthorblockN{Satvik Garg\textsuperscript{1}, Shivam Parikh\textsuperscript{2}, Somya Garg\textsuperscript{3}}
\IEEEauthorblockA{\textsuperscript{1} Hajim School of Engineering and Applied Sciences, University of Rochester, NY, USA\\
\textsuperscript{2} College of Engineering and Applied Sciences, University at Albany, State University of New York, Albany, NY, USA\\
\textsuperscript{3} Deloitte LLC, NY, USA\\
Email: \{sgarg11@ur.rochester.edu, sparikh@albany.edu, somgarg@deloitte.com\}}
}

\maketitle

\begin{abstract}

Knowledge graphs (KGs) are gaining prominence in Healthcare AI, especially in drug discovery and pharmaceutical research as they provide a structured way to integrate diverse information sources, enhancing AI system interpretability. This interpretability is crucial in healthcare, where trust and transparency matter, and eXplainable AI (XAI) supports decision-making for healthcare professionals. This overview summarizes recent literature on the impact of KGs in healthcare and their role in developing explainable AI models. We cover KG workflow, including construction, relationship extraction, reasoning, and their applications in areas like Drug-Drug Interactions (DDI), Drug Target Interactions (DTI), Drug Development (DD), Adverse Drug Reactions (ADR), and bioinformatics. We emphasize the importance of making KGs more interpretable through knowledge-infused learning in healthcare. Finally, we highlight research challenges and provide insights for future directions.

\end{abstract}


\section{Introduction}

In healthcare research, the utilization of deep learning algorithms applied to large-scale healthcare data offers an unparalleled advantage \cite{garg2020prediction, garg2021mofit}, as it enables swift and accurate decision-making processes \cite{esteva2019guide}. However, the considerable complexity of these models, characterized by millions of parameters and intricate mathematical transformations, results in their opaqueness, making them challenging for human comprehension \cite{petch2022opening}. Despite attention models providing a certain level of interpretability through knowledge visualization \cite{niu2021review}, they still need to offer explanations in a format easily understandable by humans. As a result, these models are often referred to as "black-box" models \cite{sendak2020human}, which poses significant challenges in decision-making, particularly in healthcare, where the ability to interpret and explain AI-generated recommendations is crucial \cite{gaur2021semantics}. 


As the demand for interpretability and explainability in AI systems grows \cite{fernandez2023deep, sendak2020human}, the significance of comprehending the inner workings and justifications behind decisions becomes increasingly apparent \cite{akhai2023black, dwivedi2023explainable}. 
Interpretability sheds light on system workings, promoting transparency, insights, and bias detection. Conversely, explainability justifies decisions, empowering informed choices, trust, and risk reduction \cite{akhai2023black, dwivedi2023explainable}. Knowledge Graphs (KGs) offer a promising solution for integrating diverse health data, enhancing traceability, and facilitating profound interpretations, ultimately improving decision explainability \cite{ji2021survey, garg2022birds, gaur2021semantics, roy2023proknow, gaur2020explainable}.


KGs are multi-relational graphs that can effectively reason connections between various entities \cite{ji2021survey, garg2022birds}, allowing for the integration of diverse health data, including diseases, drugs, and treatments, represented by edges with distinct labels \cite{zeng2022toward}. This unique capability to integrate knowledge, not achievable in traditional pharmacologic experiments, expedites the discovery of healthcare knowledge, leading to more profound interpretations and advancements in the healthcare domain \cite{gaur2021semantics, sendak2020human}. Moreover, KGs contain traceability, wherein decision paths can be retraced back through the graph, revealing the sequence of steps and the reasoning behind each final decision. This traceability significantly enhances the explainability of the decision-making process, making it easier to comprehend and justify the outcomes produced by the AI system.

We outline the primary contributions of this paper as follows:
\begin{itemize}
     
\item An overview of model-based approaches utilizing KGs in diverse healthcare applications. 

\item Exploration of knowledge-infused learning (K-iL) and its application in healthcare for XAI.

\item Elaboration on integrating KGs in healthcare, presenting a brief workflow containing KG construction, relation extraction techniques, and reasoning methods for eXplainable AI. This workflow provides a beginner-friendly guide for individuals entering this field.

\item Overview of state-of-the-art healthcare knowledge graphs (HKGs) and concise descriptions of their advancements.

\item Discussion of open research problems on KGs in Healthcare. Additionally, the paper addresses existing research gaps, providing valuable insights to guide future investigations.

\end{itemize}

The rest of this paper is structured as follows: Section 2 presents a comprehensive overview of the KG workflow. Section 3 touches upon knowledge-infused learning (K-iL), while Section 4 delves into a concise review of HKG's applications. Section 5 summarizes the techniques discussed and sheds light on emerging research issues and challenges. Finally, Section 6 concludes our discussion.

\begin{table*}[htbp]
\centering
\caption{Healthcare Knowledge Bases}
\label{tab:healthcare-kg}
\begin{tabular}{p{3cm} p{12cm}}
\hline
Knowledge Graph & Description \\
\hline
PrimeKG \cite {chandak2023building} & A precision medicine KG integrating 20 biomedical resources to describe 17K+ diseases with 4M+ relationships across ten biological scales, complemented by clinical guideline text descriptions. \\
CKG \cite{santos2020clinical} & A powerful open-source platform with 16M+ nodes, 220M+ relationships, and advanced algorithms for precision medicine decision-making in proteomics workflows. \\
CTKG \cite{chen2021ctkg} & Derived from ClinicalTrials.gov data, containing 1.5M+ nodes and 3.6M+ triplets, representing medical entities and their relationships in clinical trials. \\
MSI \cite{ruiz2021identification} & An interactome network integrating disease-perturbed proteins, drug targets, and biological functions for disease treatment explanation and drug-disease interaction prediction. Includes 1.6K+ drugs, 840 diseases, 17K+ proteins, and 9.7K+ biological functions. \\
Hetionet \cite{himmelstein2015heterogeneous} & A heterogeneous network with 40K+ nodes and 1.6M+ edges, predicting associations between protein-coding genes and 29 complex diseases. \\
iBKH \cite{su2021ibkh} & An integrative biomedical knowledge hub with 2M+ entities and 4.8M+ relations from diverse biomedical data resources. \\
DRKG \cite{ioannidis2020drkg} & A comprehensive biological knowledge graph with 97K+ entities and 5.8M+ triplets, relating human genes, compounds, biological processes, drug side effects, diseases, and symptoms, including Covid-19 information. \\
COVID-KG \cite{wang2021covid} & A knowledge discovery framework extracting fine-grained multimedia knowledge from scientific literature for drug repurposing in COVID-19. Includes 50K+ Gene nodes, 10.5K+ disease nodes, 5.7K+ chemical nodes, and 535 organism nodes, with 133 relation types. \\
KGHC \cite{li2020kghc} & A concise hepatocellular carcinoma knowledge graph (13K+ triples) for healthcare professionals, utilizing Neo4j for network analysis. \\
KG-COVID-19 \cite{reese2021kg} & A comprehensive knowledge graph with 377K+ nodes and 21M+ edges from 13 sources, supporting flexible remixing of subgraphs for specific areas and graph-based machine learning. \\
repoDB \cite{brown2017standard} & A comprehensive gold standard database for drug repositioning with 1.5K+ drugs and 2K+ diseases, including both approved (true positives) and disapproved (true negatives) drugs. \\
DrugBank \cite {wishart2006drugbank} & A unique bioinformatics/cheminformatics resource combining detailed drug and drug target information with 4100+ drug entries, 14,000+ linked protein sequences, and extensive drug research and pharmaceutical education applications. \\
PharmKG \cite{zheng2021pharmkg} & A biomedical KG with more than 500K interconnections between genes, drugs, and diseases, used mainly for data mining. \\
\hline
\end{tabular}
\end{table*}

\section{Knowledge Graph Workflow}

A Knowledge Graph (KG) constitutes a data representation framework based on graphs, employing triplets \emph{(e1, r, e2)} to signify knowledge through entities \textit{(e1 and e2)} that are interconnected through relationships \textit{(r)} \cite{garg2022birds, ji2021survey}. In the domain of healthcare, KGs can offer an interpretable representation of medical concepts, thereby facilitating context-aware insights and augmenting various aspects of clinical research, including healthcare delivery and decision-making \cite{santos2022knowledge, chandak2023building, chen2021ctkg}. Notably, healthcare-specific KGs are domain-specific, focusing on medical concepts like drugs and diseases.

\textit{How do knowledge graphs (KGs) enhance explainable AI models in healthcare?}
The graph-based architecture of KG's supplies them with rich knowledge into explainable AI models, where the relation triples play a critical role in influencing the credibility of specific medical information and contributing to the explainability of detection outcomes \cite{rajabi2022knowledge, rajabi2021towards, heilig2022refining}. Through effective data visualization and the preservation of relationships between interconnected information, KG's provide profound insights and elevate the comprehension of intricate medical concepts among human users \cite{wu2022knowledge}. Furthermore, the KGs inherently support the analysis of inference and causation, facilitating the exploration of causal relationships within the medical context \cite{jaimini2022causalkg, lyu2023causal}. Table~\ref{tab:healthcare-kg} presents an overview of healthcare knowledge-based, including KGs and repositories, designed to capture and integrate diverse biomedical data resources. In the following subsections, the workflow of KGs in XAI models for healthcare is outlined, including KG construction, feature extraction, and knowledge reasoning.

\subsection{Knowledge Graph Construction}  The construction of KGs involves a series of crucial steps, encompassing the definition of objectives, data gathering, transformation using specialized tools, ontology-based entity mapping, inference of missing links, and ongoing updates to enhance accuracy \cite{li2020real}. The healthcare KG integrates various concepts, such as drugs and symptoms, and establishes hierarchical and associative relationships between these entities. Furthermore, mappings to external terminologies within knowledge-based systems are included \cite{wishart2006drugbank, kanehisa2010kegg}.
The existing literature covers pivotal aspects of KG construction, including named entity recognition \cite{al2020named, jin2019named}, entity linking \cite{shi2023knowledge}, KG completion \cite{lan2021path, chen2020knowledge}, and relation extraction \cite{chen2020review}. Large language models (LLMs) have also emerged as valuable and unified tools for facilitating the construction of KGs \cite{ye2022generative, bi2023codekgc}.
To achieve a more comprehensive graph and avoid redundancy in constructing and curating KGs, researchers employ HKG's fusion to merge multiple KGs effectively \cite{youn2022knowledge, su2023biomedical}. This streamlining process mitigates the need for repeatedly creating KGs from scratch.


\subsection{Knowledge Feature Extraction}   
 

Various knowledge graph embedding (KGE) techniques exist for extracting characteristics from KGs across domains \cite{mohamed2021biological, ji2021survey}. These models learn embeddings, representing nodes and edges as low-dimensional vectors, typically with low computational demands. These embeddings are seamlessly integrated into deep neural networks and fine-tuned for specific tasks \cite{zhu2023rdkg, zeng2022toward}. For example, KGE methods find applications in drug discovery, scoring drugs based on their relationships with diseases in the KG \cite{wang2022kg, joshi2022knowledge}. These methods are valuable for drug repurposing, including for COVID-19 research \cite{rajabi2021towards}. KGs also aid in relation extraction, helping extract semantic relationships between entities, such as diseases and diagnoses or treatments \cite{gong2021smr}.

\subsection{Knowledge Graph Reasoning}

Knowledge Graphs (KGs) enable accurate predictions and inference of new facts based on existing knowledge, benefiting clinicians seeking insights from data. Transforming electronic health record (EHR) data into KGs helps identify critical clinical discoveries, enhancing understanding of medical relationships and patterns \cite{chen2020review, shang2021ehr}. Symbolic logic models, known for interpretability, mine logical rules from existing knowledge through techniques like inductive logic programming (ILP), association rule mining, and Markov logic networks (MLN). These rules improve generalization and performance in healthcare KG reasoning when combined with KG embeddings \cite{muggleton1994inductive, alshahrani2017neuro, zhu2022neural}. Human-in-the-loop techniques validate and enhance KGs, ensuring accuracy and explainability in healthcare applications. This collaboration between experts and AI systems enhances trust and reliability in AI-driven reasoning \cite{sheng2019dockg, zhang2020hkgb}.

\section{Knowledge Infused Learning}

The recent progress in deep learning language models owes its success to their adeptness in employing self-supervised objectives on extensive sets of unlabeled data \cite{liu2019roberta}. These models excel in acquiring distributional semantics and establishing connections between phrases in textual data. Nevertheless, when it comes to generating explanations and interpretations, a noteworthy challenge emerges due to the inherent limitation of internal layers or mechanisms of distributional semantics, which need domain-specific knowledge \cite{liu2020k}.

Consider a simple question-answering (QA) case where the context mentions concerns about alcoholics getting ill under lockdown. The question is whether the person has an addiction, and the answer predicted by a pre-trained language model is \emph{"Yes."} In contrast, a more complex QA scenario involves a context where the individual is discussing their experiences with depression and the challenges of accessing mental health services during the pandemic. The question is whether the individual suffers from a mental health condition such as depression, and the answer predicted by the same model is \emph{"Yes"} (although the accurate answer should be \emph{"NO"}) \cite{gaur2021semantics}.


The importance of integrating domain knowledge into deep neural models to enhance predictive capabilities and explanatory abilities is underscored. This has led to a surge in research efforts exploring methodologies to incorporate knowledge graphs (KGs) or domain rules into deep neural models for various applications \cite{gaur2020explainable, roy2023process, jaimini2022causalkg}. These approaches include integrating external knowledge as contextual information or features, infusing KG embeddings (KGEs) into neural model layers for explainability, incorporating contextual representations from relevant KG subgraphs or paths, adapting attention mechanisms using KGs, and utilizing domain-specific KGs through techniques like Integer Linear Programming (ILP) for addressing complex natural language inference tasks \cite{yao2019kg, gupta2022learning, wang2019explainable, yang2017leveraging, choi2019graph, khashabi2016question}.

\textit{What are the critical differences between explainability and interpretability through knowledge-infusion?}

Explainability pertains to the capacity of an AI system to furnish clear and coherent explanations by unraveling its decision-making processes. Expressly, explainability assumes paramount importance in healthcare as it empowers medical experts to grasp the rationale behind a particular AI-generated recommendation or outcome. Conversely, interpretability is concerned with enhancing the accessibility and understandability of the internal workings of AI models for human users. This interpretability entails simplifying intricate algorithms and model architectures so medical professionals can readily comprehend the underlying logic and mechanisms driving the AI system's behavior. In short, explainability addresses why a particular prediction has been made; in contrast, interpretability revolves around the ability to comprehend the patterns learned or knowledge acquired by the AI system. Notably, any explainable system must inherently be interpretable, but the reverse does not necessarily hold true \cite{rajabi2022knowledge}.

\section{Applications}

\begin{table*}[ht]
\centering
\small 
\caption{An overview of the recent literature utilizing KG's in healthcare applications, Drug-Drug
Interaction (DDI), Drug-Target Interaction (DTI), Adverse Drug Reaction (ADR), Drug-Repurposing (DR), Drug-Design (DD). BCE: Binary Cross Entropy, CE: Cross Entropy, PR: Pointwise Ranking, FCS: Frequent Consecutive Subsequence, PCA: Principal Component Analysis, CCE: Categorical Cross Entropy, DNN: Deep Neural Network, MLP: Multiple Layer Perceptron}
\label{tab:models}
\begin{tabular}{|p{1.4cm}|p{1.5cm}|p{0.8cm}|p{4.7cm}|p{1.5cm}|p{3.5cm}|p{1.7cm}|}
\hline
\textbf{Application} & \textbf{Model} & \textbf{Year} & \textbf{Model Components} & \textbf{Loss Function} & \textbf{Data Sources} & \textbf{Platform} \\
\hline
DDI & DDKG \cite{su2022attention} & 2022 & BiLSTM \cite{graves2013hybrid}, Attention \cite{niu2021review} & BCE & KEGG-drug \cite{kanehisa2000kegg}, OGB-biokg \cite{hu2020open} & TensorFlow \\
DDI & AMDE \cite{pang2022amde} & 2022 & FCS, MPAN \cite{withnall2020building}, Transformer \cite{niu2021review} & BCE & Drugbank \cite{wishart2006drugbank} & PyTorch \\
DDI & SUMGNN \cite{yu2021sumgnn} & 2021 & Attention \cite{niu2021review}, GNN \cite{zhou2020graph} & CE, BCE & DrugBank \cite{wishart2006drugbank}, HetioNet \cite {himmelstein2015heterogeneous}, TWOSIDES \cite{tatonetti2012data} & PyTorch \\
DDI & AAE \cite {dai2021drug} & 2021 & ComplEx  \cite{trouillon2017complex}, SimplE \cite{kazemi2018simple}, RotatE \cite{sun2019rotate}, WAA \cite{dai2020wasserstein} & BCE & DeepDDI \cite {ryu2018deep} (DrugBank \cite{wishart2006drugbank}), Decagon \cite{zitnik2018modeling} (TWOSIDES \cite{tatonetti2012data}) & - \\
DDI, DTI & MHRW2Vec-TBAN \cite{zhang2021discovering} & 2021 & SANA \cite{sun2019rotate}, ComplEx \cite{trouillon2017complex}, Metropolis-Hasting Random Walk (MHRW) \cite{zhang2021discovering}, Word2Vec \cite{church2017word2vec}, TextCNNBiLSTM-Attention Network (TBAN) \cite{zhang2021discovering} & PR & KEGG \cite{kanehisa2010kegg}, DrugBank \cite{wishart2006drugbank}, PharmGKB \cite{zheng2021pharmkg} & TensorFlow \\
DTI & KG-DTI \cite{wang2022kg} & 2022 & DistMult \cite{yang2015embedding},  Conv-Conv module \cite{wang2022kg}  & BCE & - & - \\
DTI & KGE-NFM \cite{ye2021unified} & 2021 & DistMult \cite{yang2015embedding},  PCA, Neural Factorization Machine (NFM) \cite{he2017neural} & BCE & BioKG \cite{walsh2020biokg}, Yamanishi08 \cite{yamanishi2008prediction} & TensorFlow \\
DD & Ranjan et al. \cite{ranjan2022generating} & 2022 & GGNN \cite{beck2018graph},  GCN \cite{zhang2019graph}& - & MOSES \cite{polykovskiy2020molecular}, DAVIS \cite{davis2011comprehensive}& PyTorch \\
DD & Li et al. \cite{li2022prediction} & 2022 & Random Forest, DNN, GCN \cite{zhang2019graph}, CMPNN \cite{song2020communicative}& BCE & Pistachio \cite{nextmovesoftwareNextMoveSoftware} & TensorFlow \\
DR & KANO \cite{fang2023knowledge} & 2023 & OWL2Vec \cite{chen2021owl2vec}, CMPNN \cite{song2020communicative}, MLP, Self-Attention \cite{niu2021review}, GRU \cite{dey2017gate} & NT-Xent \cite{chen2020simple}, BCE & ElementKG \cite{fang2023knowledge}, ZINC15 \cite{sterling2015zinc}& PyTorch \\
DR & DrugRep-KG \cite{ghorbanali2023drugrep} & 2023 & Word2Vec (CBOW) \cite{church2017word2vec}, Logistic Regression Classifier& BCE & DrugBank \cite{wishart2006drugbank}, repoDB \cite{brown2017standard}, SIDER \cite{kuhn2016sider}, DisGeNET \cite{pinero2020disgenet} & Scikit Learn \\
DR & KG-Predict \cite{gao2022kg} & 2022 & CompGCN \cite{vashishth2019composition}, InteractE \cite{vashishth2020interacte} & CCE & DrugBank \cite{wishart2006drugbank}, TreatKB \cite{wang2018disease}& PyTorch \\
ADR & KGDNN \cite{joshi2022knowledge} & 2022 & Node2Vec \cite{grover2016node2vec}, Word2Vec (CBOW) \cite{church2017word2vec}, DNN & BCE & SIDER \cite{kuhn2016sider}, Drugbank \cite{wishart2006drugbank}, Drug-Path \cite{zeng2015drug}, DGIdb \cite{griffith2013dgidb} & Keras \\
\hline
\end{tabular}
\end{table*}

Table~\ref{tab:models} illustrates the extensive utilization of KGs across diverse domains within drug research, showcasing their versatility and effectiveness.

\begin{itemize}

 \item  Drug-Drug Interactions (DDIs):  KGs serve as a valuable resource for modeling and analyzing interactions between different drugs. Through the representation of drugs and their properties as nodes and edges within the graph, KGs facilitate the identification of potential drug-drug interactions and the assessment of their impact on patient safety and efficacy \cite{su2022attention, pang2022amde, yu2021sumgnn}.
 
 \item Adverse Drug Reactions (ADRs): KG's offer valuable insights into adverse reactions associated with specific drugs. By integrating data on drug properties, patient profiles, and reported adverse events, KGs enable the detection of patterns and potential risks associated with particular medications \cite{joshi2022knowledge}.
 \item  Drug-Target Interactions: KGs prove instrumental in mapping relationships between drugs and their corresponding molecular targets. This molecular mapping enhances drug discovery endeavors and explains the more profound understanding of the mechanisms underlying drug action, contributing to developing more targeted and effective therapeutic interventions \cite{ye2021unified, wang2022kg, zhang2021discovering}.
 \item  Drug Development (DD): KGs significantly stream the drug development process. Through the integration of data from diverse sources, including clinical trials, scientific literature, and molecular databases, KGs offer valuable support for decision-making in drug development \cite{ranjan2022generating, li2022prediction}, optimization, and repurposing efforts \cite{fang2023knowledge, ghorbanali2023drugrep}.
 \item  Bioinformatics Research: In bioinformatics, KGs are potent tools for integrating and analyzing biological data, encompassing protein-protein interactions, gene-disease associations, and pathway networks. These comprehensive graphs empower researchers to look at critical insights into complex biological processes and identify potential targets for further investigation \cite{dimitrov2022comparison, feng2023genomickb}.
 
\end{itemize}

\section{Open Research Challenges}
\begin{itemize}
    
\item  Open World Assumption: Constructing KGs under the open-world assumption involves building KGs without relying on pre-defined schemas or exhaustive entity and relation normalization \cite{li2022oerl, lu2022open}. By adopting this open-world assumption, researchers aim to increase the coverage of the KG to capture a broader range of information and relationships. However, while open-world KGs offer expanded coverage, ensuring the quality of the extracted knowledge becomes a critical research challenge, particularly for creating explainable and trustworthy HKGs \cite{wang2021covid, he2020research} 

\item  Knowledge Integration: Healthcare KG integration, or healthcare KG fusion, involves combining two or more HKGs into a single, more comprehensive graph \cite{youn2022knowledge}. However, different HKGs employ diverse terminologies, schemas, and data formats, making seamless integration difficult. To overcome this challenge, developing robust techniques for ontology matching becomes crucial \cite{he2022machine}. Additionally, reconciling the different graph structures through schema alignment is another open research challenge. The varying organization of data in different HKGs requires efficient algorithms to establish connections and relationships between entities \cite{hu2021integrated, bachman2018famplex}. Conflict resolution methods are also necessary to adjust conflicting information from different HKGs, ensuring the integrity and reliability of the merged graph \cite{ma2023tecre}. 

\item  Model Complexity: Achieving a good balance between the explainability and performance of an eXplainable AI (XAI) model becomes challenging due to the substantial diversity and complexity of the available data \cite{han2021challenges}. Moreover, the traversal operations on large-scale KGs with numerous vertices and high outdegree pose a significant obstacle in the knowledge discovery process. Navigating the abundance of options to identify relevant paths and closest facts can be difficult. Moreover, the complexity of the models can lead to learning biases, particularly concerning certain types of biomedical data, thereby impacting the quality of results and their explainability. Therefore, addressing these learning flaws in critical healthcare applications may take precedence over emphasizing explainability.

\item  Benchmarking and Validation: Due to the complexity of constructing KG's and the diversity of data sources in healthcare \cite{he2020research}, validation methods become necessary to verify the correctness and consistency of the information within the graph.
Validating the knowledge graph helps guarantee that the explanations derived from it are trustworthy and accurate \cite{bean2017knowledge}, thereby enhancing the overall reliability and trustworthiness of XAI systems in various health-based real-world applications \cite{paulheim2017knowledge}. 

\item User-Friendly Explanations:  Many people need more technical knowledge to assess the quality of AI decisions, making it difficult to trust or interpret AI-driven outcomes \cite{de2022perils}. The main challenge in eXplainable AI (XAI) is providing user-friendly explanations that can be easily understood by users, particularly by individuals without expertise in AI. KG offers relief to this challenge by presenting visualizations and semantic representations of concepts \cite{wu2022knowledge}. However, further research is needed to optimize these methods and make AI explanations more accessible and comprehensible to a broader audience. Addressing this challenge is essential for fostering transparency and trust in AI systems across diverse user groups.

\item  Privacy and Security: Healthcare data often contain sensitive and private information, which makes data privacy and security a top concern when constructing and utilizing knowledge graphs. Developing robust methods to anonymize and protect patient data while maintaining the utility of the graph for research and clinical purposes is essential \cite{sharma2022privacy, kim2021semantically}. Moreover, applying KG's in healthcare raises ethical and regulatory issues related to data ownership, consent, and responsible use of AI-driven insights. Addressing these concerns is crucial for gaining acceptance and trust from patients, clinicians, and healthcare institutions.


\end{itemize}

\section{Conclusion}

Knowledge graphs (KGs) have appeared to be transformative in the healthcare domain, significantly enhancing the interpretability and explainability of AI-driven processes. By leveraging KGs, healthcare professionals can access clear explanations for AI decisions, fostering trust and informed decision-making. In this study, KGs find diverse applications in healthcare, addressing drug interactions, adverse reactions, drug development, and bioinformatics research. Moreover, the concept of Knowledge-infused Learning, which involves injecting background knowledge, is discussed, which could be a game-changer and highly beneficial for general health-based applications like QA systems. They bridge the gap between healthcare experts and AI models, advancing research and patient care.
Nevertheless, challenges persist in ensuring knowledge quality, model reasoning complexity, integration of diverse KGs, and user-friendly explanations. Overcoming these hurdles will lead to more reliable and trustworthy AI-driven healthcare applications, ultimately improving patient outcomes. Additionally, safeguarding the privacy and security of healthcare data remains vital to ensure widespread acceptance and trust in AI systems within the healthcare domain. In summary, healthcare knowledge graphs hold immense potential to shape a more interpretable and transparent healthcare ecosystem, promoting evidence-based practices to benefit patients and healthcare providers.

\bibliographystyle{plain} 
\bibliography{satvi}

\end{document}